# DULA and DEBA: Differentiable Ergonomic Risk Models for Postural Assessment and Optimization in Ergonomically Intelligent pHRI

Amir Yazdani[1], Roya Sabbagh Novin[1], Andrew Merryweather[1], and Tucker Hermans[2]

***Abstract*— Ergonomics and human comfort are essential concerns in physical human-robot interaction applications. Defining an accurate and easy-to-use ergonomic assessment model stands as an important step in providing feedback for postural correction to improve operator health and comfort. Common practical methods in the area suffer from inaccurate ergonomics models in performing postural optimization. In order to retain assessment quality, while improving computational considerations, we propose a novel framework for postural assessment and optimization for ergonomically intelligent physical human-robot interaction. We introduce DULA and DEBA, differentiable and continuous ergonomics models learned to replicate the popular and scientifically validated RULA and REBA assessments with more than 99% accuracy. We show that DULA and DEBA provide assessment comparable to RULA and REBA while providing computational benefits when being used in postural optimization. We evaluate our framework through human and simulation experiments. We highlight DULA and DEBA's strength in a demonstration of postural optimization for a simulated pHRI task.**

## I. Introduction and Motivation

Improving workplace ergonomics and reducing the risk of work-related musculoskeletal disorders (WMSDs) has been the focus of researchers for many years [1,2]. High rates of injuries in industry [3] further motivates these studies. The industry 4.0 and 5.0 initiatives [4,5] and the development of collaborative workplaces where smart agents (e.g., cobots [6]) physically collaborate with humans to perform different types of tasks, highlights the growing importance of human comfort and ergonomics in physical human-robot interaction (pHRI). Moreover, the fact that smart agents share a physical interaction point with the operator provides the opportunity to develop novel technologies for those agents to consider human comfort in their behaviors.

A common and practical solution to improve workplace ergonomics is to perform risk assessments and analyze human comfort during task execution. Ergonomists usually focus their analysis on the worst posture achieved during the task by taking measurements of the worker's posture onsite or from recorded images or videos. They normally use risk assessment tools [7] to assess the task and provide intervention and improvement in forms of user training [8], workstation modification [9] or task rotation [10]. However, in most cases, it is not possible to have an ergonomist present during the task to perform a comprehensive risk assessment, which results in missing events or only partial understanding of awkward postures. In addition, these sampling strategies lead to failures in detecting the most hazardous combinations of force and posture, increasing the inaccuracy of an assessment. Finally, although postural correction by ergonomists generally helps reduce the risk of injuries, in many cases, the corrections are sub-optimal or tailored to specific task requirements, which might not generalize across task variability.

[1]Department of Mechanical Eng. and Robotics Center, University of Utah, Salt Lake City, UT, USA, amir.yazdani@utah.edu. [2]School of Computing and Robotics Center, University of Utah, Salt Lake City, UT, USA, and NVIDIA, Seattle, WA, USA. Research reported in this publication was supported in part by DARPA under grant N66001-19-2-4035 and the National Institute for Occupational Safety and Health under award number T420H008414-10.

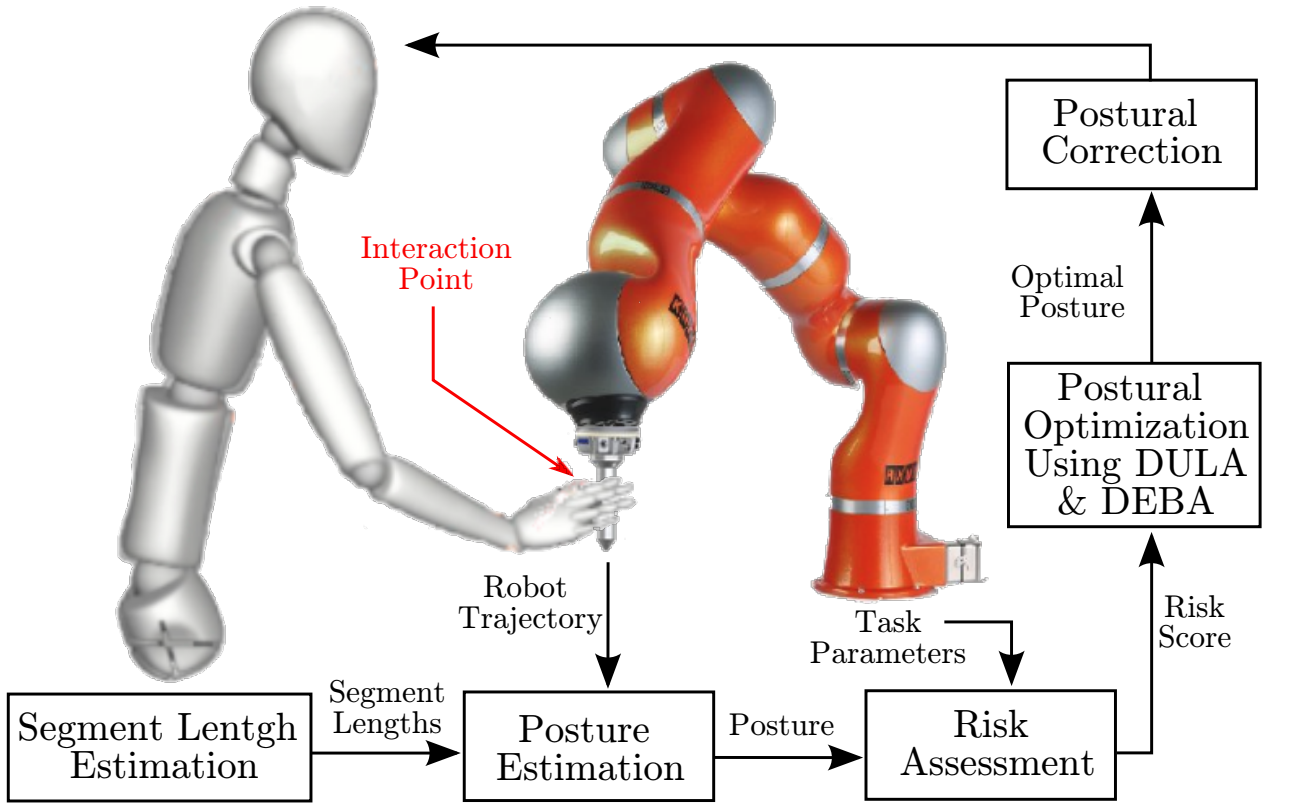

**Fig. 1:** The framework for ergonomically intelligent pHRI, adapted from [11].

To provide a solution for the above issues, we previously introduced a framework for an *ergonomically intelligent teleoperation system* in [11], which includes smart and autonomous posture estimation, postural ergonomics assessment, and postural optimization. Figure 1 visualizes our proposed framework. As a first step to realizing this system, we investigated a new approach for posture estimation in teleoperation relying solely on sensing from the leader robot [12]. Although, our implementation focused on teleoperation, it can be extended to other pHRI tasks with minor modifications. In this paper, we focus on ergonomics assessment and postural optimization in the context of pHRI.

pHRI covers a vast number of applications. To help present our framework, we categorize pHRI tasks into *direct physical interaction* with the robot (e.g., robotic physical therapy [13]), *co-manipulation* [14], *assistive holding* [15], *handover* [16], and *teleoperation* [17]. We split teleoperation into three main types: *goal-constrained teleoperation* (e.g., pick and place with a desired pose for the object), *path-constrained teleoperation* (e.g., turning a valve where the path to follow is constrained based on the diameter of the valve), and *trajectory-constrained teleoperation* (e.g., weld-ing or sealing where the operator should follow a reference travel velocity in addition to a defined path to perform a uniform and good quality weld or seal). Figure 2 visualizes the different types of pHRI applications.

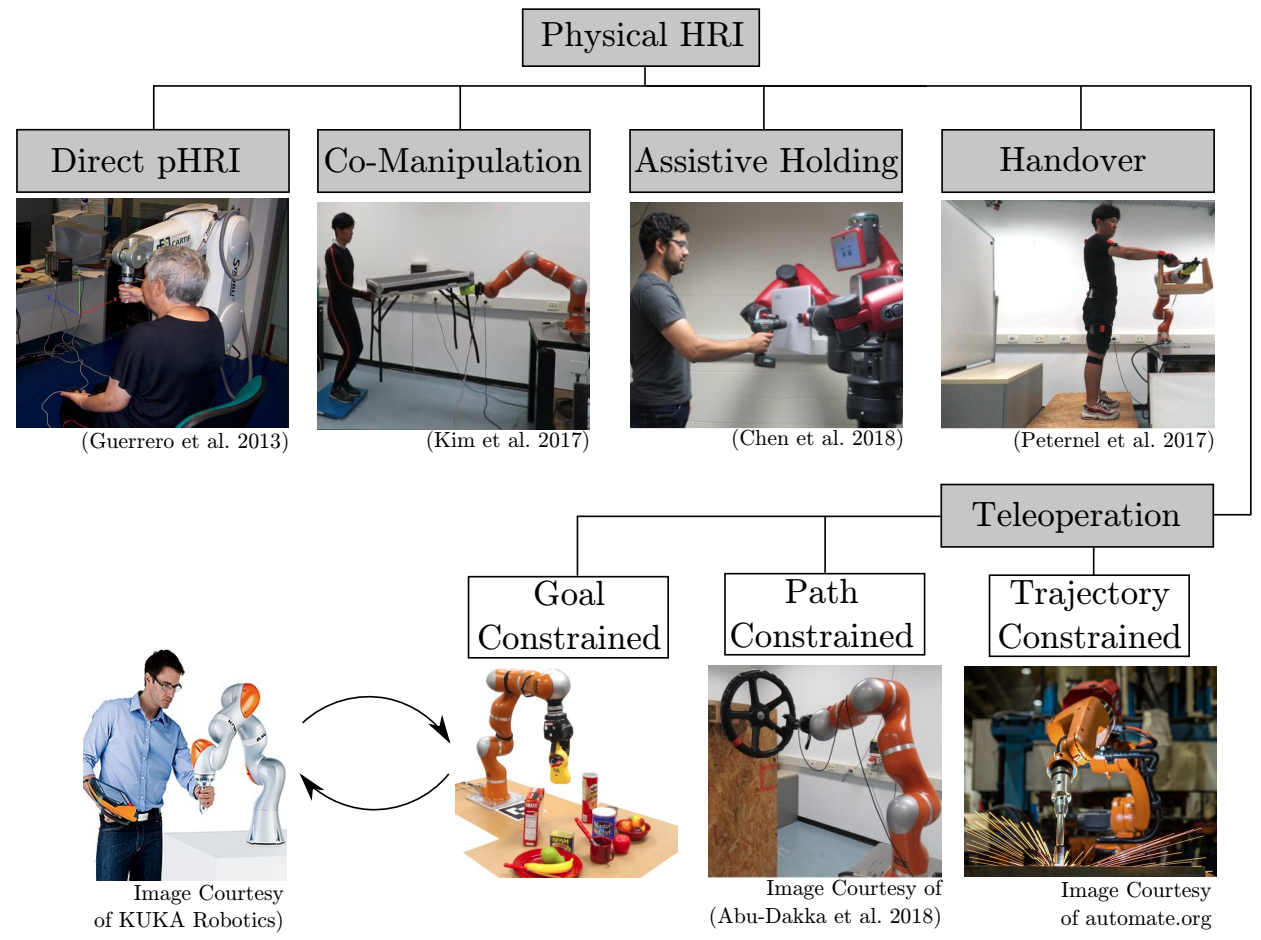

**Fig. 2:** Categories of physical HRI applications.

*Autonomous Postural Optimization* has received substantial research attention using various technologies around humans such as collaborative robots [18], smart personal trainers [19], and VR systems [20]. In these systems, the interacting agent should consider human comfort and ergonomics in its behaviour and motion planning [11]. For example, in co-manipulation the collaborative robot should account for satisfying ergonomics and comfort of the human during the collaboration when generating controls. Ergonomics and comfort can be included in postural optimization in two ways: (1) as a constraint, and (2) as a part of the objective function. In the above example, the robot can behave in such a way that avoids high-risk and awkward postures (comfort as a constraint) or minimizes the risk of musculoskeletal disorders for the whole duration of the task (comfort as a part of the objective function). The literature in postural optimization in pHRI and teleoperation is minimal. Table I summarizes the relevant literature in both areas. Most works use gradient-based optimization techniques for postural optimization [14–18,21–23], while a couple opt for gradient-free methods [24,25]. Although gradient-free approaches are more time and resource consuming, they can work with non-differentiable objective functions. In contrast, gradient-based approaches are usually faster, but require the costs and constraints to be continuous and differentiable.

Developing a model of human comfort lies at the heart of effective postural optimization. Ergonomists have provided simple models which are easier for human experts to calculate by hand and as such are common in practice. These models include the NASA TLX [26], RULA [27], REBA [28], strain index [29], and ACGIH TLV [30]. Importantly, these models are supported by extensive human subject studies that validate their effectiveness on reducing ergonomic risk factors [31]. pHRI researchers have also proposed computational models for human comfort, including peripersonal space [15], muscle fatigue [21], joint overloading [14], and muscular comfort [15].

Among all ergonomics risk assessment tools, RULA [27] and REBA [28] depend most directly on human posture and provide quantitative scores, making them good choices for postural optimization applications. RULA targets the upper body, while REBA assesses the whole body posture of the human. Both models take as input the joint angles of the human and output discrete, integer scores where the higher number indicates the higher risk of WMSDs, i.e. lower human comfort [32]. The interpretation of each score can be found in [27,28]. Van der Spaa et al. provided the only study of directly using risk assessment tools in derivative-free postural optimization [25].

The discrete, step-function like output scores in RULA and REBA [23] prevent its use in gradient-based postural optimization, as the gradient is either 0 or undefined. Based on our experience, using the risk assessment models directly in gradient-free optimization is also time-expensive, as the plateaus often prevent progress toward the optimal global solution. Thus, researchers in pHRI often use computational approximations of ergonomic assessment models to enable gradient-based postural optimizations. As Table I shows, quadratic approximations [16,17,23] make up the standard approach in the literature. However, these approximations deviate far from the scientifically validated assessments, causing uncertainty about their reliability in providing ergonomic assessment and improvement.

In contrast, this paper presents a continuous and differentiable, tight approximation of the fully validated RULA and REBA assessments tools. This enables us to improve both gradient-based and gradient-free optimization using the continuous models. We propose two continuous and differentiable ergonomics models for postural optimization, *Differentiable Upper Limb Assessment* (DULA) and *Differentiable Entire Body Assessment* (DEBA) approximating the RULA and REBA survey tools respectively. We build DULA and DEBA as neural network regression models, which we train to predict the same output scores as RULA and REBA, from associated human postures. However, DULA and DEBA output continuous, not discrete, values across the assessment range enabling smoother optimization landscapes compared to the non-differentiable models. Furthermore, they provide the gradient of the risk with respect to each joint enabling efficient use in optimization. We show that both models achieve over 98% agreement with RULA and REBA on large scale synthetic datasets. We then investigate the utility of DULA and DEBA to improve both gradient-based and gradient-free postural optimization in pHRI and teleoperation. We provide our models for others use at https://sites.google.com/view/differentiable-ergo.

We organize the remainder of the paper as follows. Section II introduces our framework for postural optimization in pHRI. Section III defines the structure, training, and data generation for DULA and DEBA. We define our ergonomic assessment and optimization experiments in Sec IV and analyze our results in Sec V. We conclude in Sec. VI.

| Ergonomics Model | | Analytical Models | Learned Models | Risk Assessment Tools | |
|---|---|---|---|---|---|
| | | | | Quadratic Approximation | No Approximation |
| Optimization | Gradient-Based | ■ Peternel et al. [18]<br>▪■ Peternel et al. [21]<br>□ Chen et al. [15]<br>◆ Peternel et al. [22]<br>■ Kim et al. [14] | This work | ◇Rahal et al. [17]<br>▪ Busch et al. [23]<br>▪ Busch et al. [16] | Infeasible for current assessment tools |
| | Gradient-Free | | | | ■ Van der Spaa et al. [25] |
| Non-Optimization | | | | | □ Shafti et al. [33] |
| Legend | | □ pHRI: Assistive Holding, ▪ pHRI: Handover, ■ pHRI: Co-Manipulation<br>◆ Teleop: Goal-Constrained with Postural Optimization By Interface Reconfiguration<br>◇ Teleop: Goal-Constrained with Online Postural Optimization | | | |

**TABLE I:** State of the art of postural optimization in pHRI and teleoperation.

## II. Postural Optimization in pHRI and Teleoperation

In this section, we introduce the problem of postural optimization in pHRI. We focus our presentation on the applications of co-manipulation and teleoperation since they contend with the largest ranges of continuous motions. However, the same problem statement can be extended to other applications with minor modifications. We present the problem formulation with a generic "comfort" objective. We formally define our proposed metrics in Section. III.

### A. Postural Optimization in Co-Manipulation

In co-manipulation, the robot helps a human to move an object (usually a heavy one) from an initial pose to a desired pose [21]. There is no specific trajectory for the object to follow; the human plans and leads the motion, and the robot follows and provides help carrying the object. Meanwhile, ergonomically intelligent pHRI seeks to find the optimal posture for the human at each instance while holding the pose and velocities at the interaction point close to the current state. Once found, the system will suggest the solution to the human. We call this *online postural optimization*. We can formalize finding the ergonomically optimal posture with the following optimization problem:

$$\begin{aligned} \mathbf{q}_t^* = \arg\max_{\mathbf{q}_t} \quad & \text{Comfort}(\mathbf{q}_t) \qquad (1) \\ \text{s.t.} \quad & ||\phi([\mathbf{q}_t;\dot{\mathbf{q}}_t],\psi) - [\mathbf{z}_t;\dot{\mathbf{z}}_t]||_\Sigma^2 < \epsilon \\ & \mathbf{q}_t \in \text{Range of Motion} \end{aligned}$$

where $\mathbf{q}_t^*$ and $\mathbf{q}_t$ are the optimal posture and current posture at time $t$, respectively, $\phi$ is the forward kinematics of the human, and $\psi$ is the human segment lengths. It is important to note that the optimal posture $\mathbf{q}_t^*$ from Eq. (1) for each time step is then suggested to the human to move towards. The human can refuse or accept the correction as much as they desire while completing the task.

### B. Postural Optimization in Teleoperation

The main distinction of teleoperation from other pHRI applications is the motion (and force) coupling between the leader and follower robots. The coupling between the two need not be a one-to-one scale and can be paused and resumed during task execution. This enables the teleoperator to disengage the leader robot in the middle of a task in many types of application, reposition their postures (and the leader robot as the result), and resume the teleoperation from a more comfortable posture [22]. The relative position trajectory of the follower robot remains unchanged; however, its velocity trajectory changes due to the pause. The motion scaling also helps keep the teleoperator's posture in a smaller comfort zone while the follower robot operates in a much wider zone. To use this feature of teleoperation in postural optimization, we customize three postural optimization problems for the three teleoperation problems shown in Fig. 2.

*1) Online Postural Optimization:* Online postural optimization in teleoperation is very similar to online postural optimization in co-manipulation. The difference is that instead of having initial and goal poses for the shared object in co-manipulation, in teleoperation, we have initial and goal poses for the end-effector of the follower robot (or the object it manipulates). Hence, the ergonomically intelligent system should find the optimal posture of the human at each instance, in which the pose and velocities of the follower robot's end-effector are not far from the current states, and suggest it to the human. The formulation in Eq. (1) works here too, where $[\mathbf{z}_t;\dot{\mathbf{z}}_t]$ are observed from the leader robot.

*2) Initial Postural Optimization:* The remote connection between the leader and follower robots makes it possible to start the teleoperation task from any initial human posture and corresponding initial posture of the leader robot without changing the follower robot. We use this feature to propose *initial postural optimization* for path-constrained and trajectory-constrained teleoperation tasks. In this type of postural optimization, the ergonomically intelligent system can get info about the path (or trajectory) of the task either by observing the human doing the task once, or using inference approaches based on priors on the task objectives (e.g. using object pose estimation techniques to understand the diameter, center, and the rotation axis of a valve to rotate [34]), then calculates the optimal initial posture for the human to start *the same task* with the same task-space motion for the follower robot. Formally we define the problem

$$\begin{aligned} \mathbf{q}_0^{h^*} = \arg\max_{\mathbf{q}_0^h} \quad & \sum_{t=0}^{T} \text{Comfort}(\mathbf{q}_t^h) \qquad (2) \\ \text{s.t.} \quad & ||\hat{\dot{\mathbf{x}}}_t^h - J^h(\mathbf{q}_t^h,\psi)\dot{\mathbf{q}}_t^h||_\Sigma^2 < \epsilon \quad \forall t \geq 0 \end{aligned}$$

where $\hat{\dot{\mathbf{x}}}_t^h$ is the recorded or inferred velocity profile of the interaction point during the task, and $J^h(\mathbf{q}_t^h, \psi)$ is the Jacobian of the human kinematic chain. Here, the superscripts $h, l, f$ refer to human, leader robot, and follower robot, respectively.

*3) Postural Optimization by Interface Reconfiguration:* An inherent feature of path-constrained teleoperation is the ability to pause the teleoperation by disengaging the leader and follower robots from each other, move the leader robot to a new position, and resume the teleoperation. This is usually done by a clutch switch placed under the user's foot. To benefit from this feature, we propose *postural optimization by interface reconfiguration*. In this case the ergonomically intelligent system suggests to pause the teleoperation and reconfigure the human posture when it detects high-risk postures during a task. When teleoperation is paused, the system calculates a new optimal posture for the human to resume the teleoperation from and continue the task.

The formulation is the same as that of initial postural optimization, Eq. (2), but only covers the time after the pause. We define $t_p$ as the time at pause in teleoperation. Then we simply change the indices for the summation in the objective and enumeration of constraints to start at $t_p$ instead of 0. Similar to the initial posture optimization, $\hat{\dot{\mathbf{x}}}_t^h$ can come from a variety of sources. This could be derived from a history of a human performing the task or inferred from task objectives (e.g. using object tracking to locate the welding seam [35] and use the prior knowledge on proper reference travel velocity to follow [36]).

## III. Differentiable Ergonomics Models

This section proposes our approach to develop differentiable comfort models for use in the objective functions in Sec. II. We achieve this by learning neural network models that replicate the output of the RULA and REBA models. By using neural networks we produce continuous, differentiable models to replace the discrete, non-differentiable RULA and REBA. We name the differentiable variants of these models *DULA* (Differentiable Upper Limb Assessment) and *DEBA* (Differentiable Entire Body Assessment). We train these models using a standard supervised learning approach. In this remainder of this section we give the details of the DULA and DEBA architectures, training the models and generating the training data.

### A. DULA and DEBA

To learn continuous and differentiable models approximating RULA and REBA, we designed fully connected neural networks for regressions. While RULA provide discrete integer scores from 1 (lowest risk of WMSDs) to 7 (highest risk of WMSDs), and REBA provide discrete integer scores from 1 (lowest risk of WMSDs) to 15 (highest risk of WMSDs), we choose to predict continuous labels for those models. If we had instead performed multi-class classification based on the discrete labels, the resulting model would be less useful for optimization as there is no natural choice of a smooth function to minimize or constrain ergonomic cost. This would negate our desire for a computationally helpful

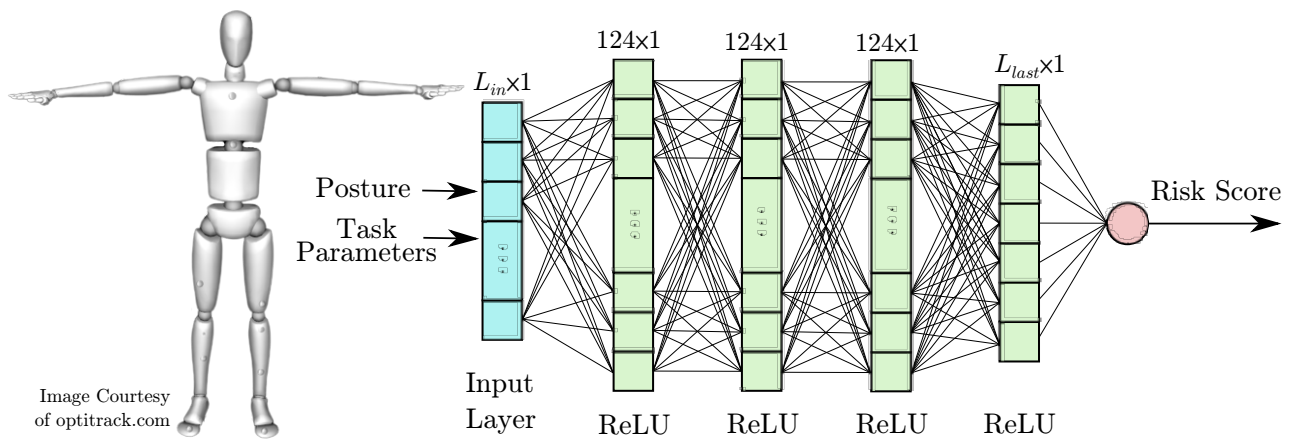

**Fig. 3:** The structures of the neural networks for learning DULA and DEBA. For DULA $L_{in}$ = 10+task parameters and $L_{last}$ = 7, where for DEBA $L_{in}$ = 11 + task parameters and $L_{last}$ = 12.

model. Hence, we perform continuous regression to the multi-class integer labels.

We calculate the loss for each forward path of the training based on the difference between the continuous score from the forward path and the integer label for the data point

$$\frac{1}{N}\sum_{i=1}^{N} ||y_i - f(\mathbf{q}_i, \theta)||_2^2 \tag{3}$$

where $y_i$ is the label, $\mathbf{q}_i$ is the input posture, $\theta$ is the task parameters, and $N$ is the number of data points.

However, in calculation of the models' accuracy, we round the continuous scores to their nearest integers, and then compare them with the integer labels from RULA and REBA.

Looking at the REBA literature [28], we observe that the *activity score* parameter is added directly to the output of Table C to calculate the final REBA score. As the information for computing the activity score is available to us a priori and remains constant during a task, we do not want to re-learn it. We thus exclude it from the learning process and design the output of the network to represent scores from 1 to 12.

The structure of the neural networks for DULA and DEBA are shown in Fig. 3. We performed 5-fold cross-validations to find the optimal network parameters, which revealed that the networks with 4 hidden layers with ReLU activation function, with 124 units for the first three layers and 7 units for DULA and 12 units for DEBA at the last hidden layer worked the best. We trained the networks using the loss function in Eq. (3) for 2000 epochs using a learning rate of 0.001. Training of the neural networks are done on a NVIDIA 1070-Ti GPU. We implemented and trained our network using a mixture of tools from PyTorch [37], scikit-learn [38], and skorch [39].

### B. Dataset for Learning DULA and DEBA

We generated two datasets of 6 million upper and full body postures each for DULA and DEBA respectively. We additionally defined task parameters based on the RULA and REBA worksheets—the frequency of the arm and body motions; type and maximum load on the arm and body; neck angle; the activity score; coupling score; and whether any legs, feet, or arms are supported. We developed scripts for automatic RULA and REBA assessment based on the posture and tasks parameters and verified it with several ergonomists. We used these scripts to label the posture datasets. Since postures with RULA labels 1, 2, 6, and 7 and DEBA labels of 1, 2, 3 are not frequent in the full range of human motion, we balanced the dataset by setting the data generation scripts to generate sample data points around the neutral posture until we have an equal number of data with each label. We split the dataset into 80% training and 20% testing sets.

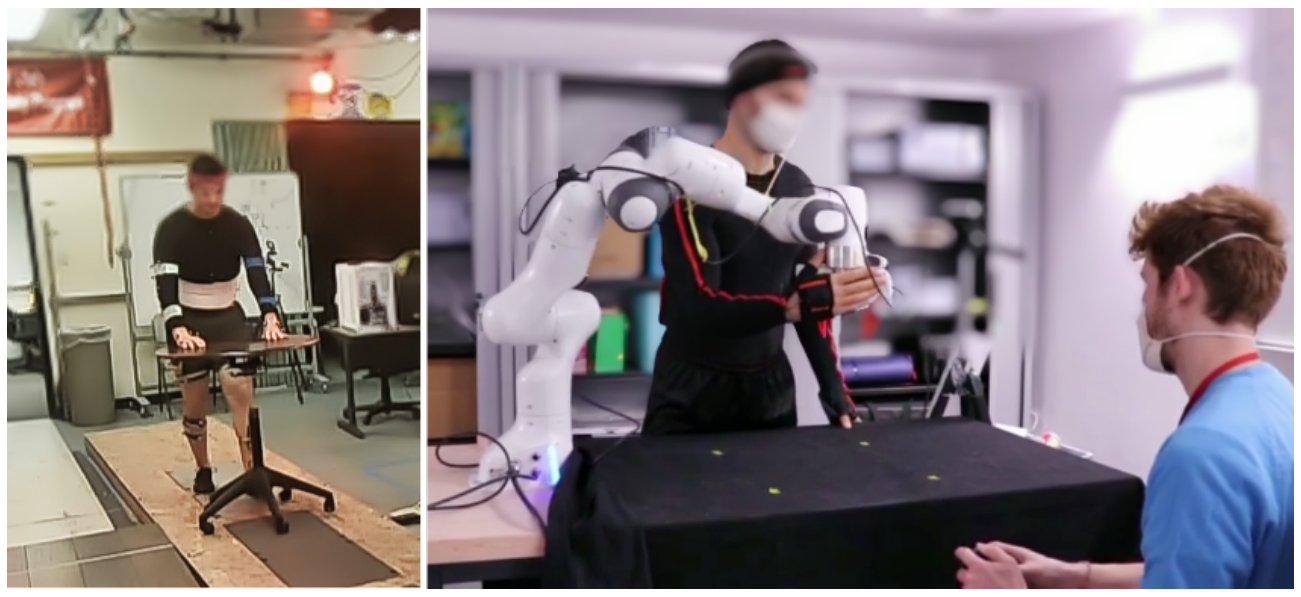

**Fig. 4:** Examples of human subject studies: (left): pushing a table, and (right) interacting with a remotely-controlled robot from [40].

## IV. Human Subject and Simulation Experiments

In this section, we share the details of the postural optimization algorithms and the simulation environment we developed to evaluate the effect of postural optimization in reducing the risk of WMSDs.

### A. Postural Assessment Using DULA and DEBA

To compare the output of DEBA to REBA, we conducted a human subject study[1] in which participants perform four common tasks in industrial environments including lifting and twisting, pulling, pushing, and reaching. We recorded their postures using an OptiTrack motion capture system and assessed them using RULA/REBA and DULA/DEBA ergonomics assessment models. We recruited 4 participants, 22-31 years old from university students.

We additionally evaluate the dataset of human motions during a pHRI subject study from Vianello et al. [40] using RULA/REBA and DULA/DEBA. In this human subject study, the human performs a 3 type of co-manipulation tasks with a Franka Emika Panda robot. The subject's posture is recorded using a Xsens MVN suit. Figure 4 shows snapshots of these two studies corresponding to trials in Fig 7.

### B. Simulated Environment For Postural Optimization

We developed an open-source simulator for postural correction using ROS. It includes a human and a KUKA LWR-4 robot (two KUKA robots for teleoperation). The simulated human behaves like a human in two ways: (1) physically controlling the interactive task, and (2) accepting or rejecting the recommended optimal postural corrections. We model this in an optimal motion planning framework with re-planning that finds a human joint trajectory that interacts with the robot to do the task while moving toward the optimal ergonomic posture. For example, it is formulated as the following for a teleoperation task:

$$\tau^{h^*}_{t\to H} = \arg\min_{\tau^{h}_{t\to H}} \sum_{t=t}^{H} ||\mathbf{x}^f_g - \mathbf{x}^f_t||^2_\Sigma + \alpha||\mathbf{q}^h_t - \mathbf{q}^{h^*}_t||^2_2 \quad (4)$$

[1]IRB number: IRB_0040280

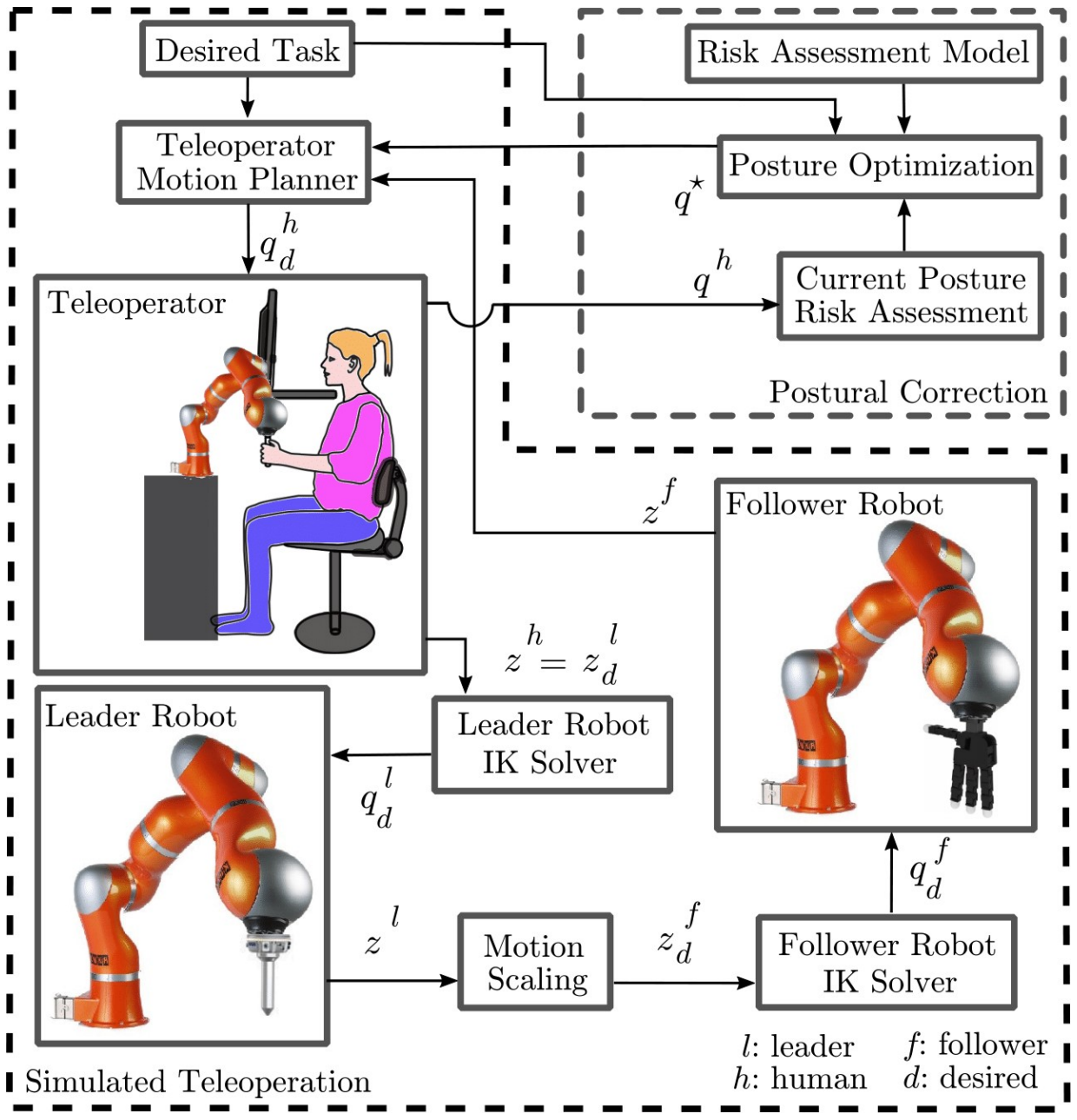

**Fig. 5:** Overview of teleoperation simulation.

where $\tau^f_{t\to H}$ is the trajectory of human posture from time $t$ to the time of the horizon $H$, $\mathbf{x}^f_g$ is the goal pose of the follower robot, and ${}^f\mathbf{x}_t$ is the pose of the follower robot at time $t$. Here, $0 \le \alpha \le 1$ is a scalar number that models the postural correction acceptance and the effort of the human operator towards applying the postural correction. Details of the motion planning for the humans and the robots are presented in Fig. 5. We note that this model is likely a simplification of true human interactive behavior. We do not advocate for its use over human subject studies. Instead, we propose it as a useful tool for systematically exploring new human safety assessment and improvement algorithms.

### C. Methods for Postural Optimization

We explore two different solvers for postural optimization, one gradient-based (SQP [41]) and one gradient-free (CEM [42]). We investigate the use of DULA/DEBA with both solvers by replacing the objective of maximizing human comfort in the problems from Sec. II with minimizing the ergonomic risk encoded by e.g. RULA or DEBA.

$$\mathbf{q}^* = \arg\min_{\mathbf{q}} \quad \mathrm{RULA/REBA/DULA/DEBA}(\mathbf{q}) \quad (5)$$

Our primary interest lies in examining to what extent the use of derivative computation in DULA/DEBA improves performance over the gradient-free optimization required in using RULA/REBA as an objective. However, as a secondary question we seek to examine if using the continuous DULA/DEBA models improves gradient-free optimization performance over the discrete RULA/REBA models.

We use the SLSQP solver from SciPy [43] for SQP. We calculate DULA and DEBA gradients using automatic differentiation in PyTorch. For CEM, we use a Gaussian sampling distribution with 10000 samples.

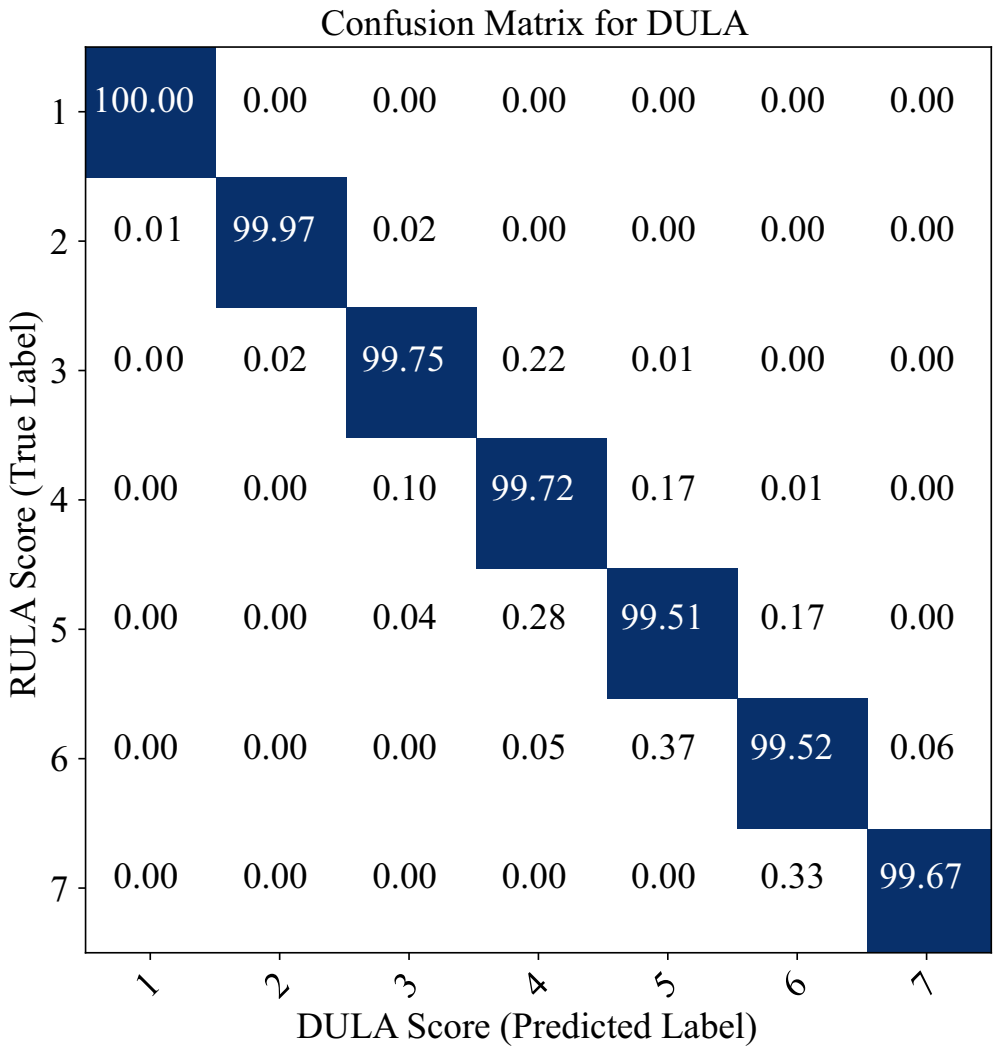

**Fig. 6:** Confusion matrix (accuracy %) for DULA vs RULA.

## V. Results and Discussion

This section showcases the accuracy results for DULA and DEBA and the effect of their use in postural optimization.

### A. DULA and DEBA Accuracy

Fig. 6 represents the confusion matrix for DULA. The trained model for DULA performs well on the test dataset and results in 99.73% accuracy. The lowest diagonal element in the confusion matrix shown in Fig. 6 for DULA is 99.38%, which shows the high accuracy of the learned model across all ranges. DEBA also has a high accuracy of 99.16% and the lowest diagonal element of the confusion matrix is 98.5%.

Training DULA took 61 hours, while DEBA took 64 hours. Note, however, that this training only has to take place once as the models can be used for all relevant ergonomic assessment tasks.

### B. Postural Assessment

Figure 7 compares the ergonomics assessment scores during representative trials from the pushing (top), and pHRI co-manipulation tasks (bottom). We compare RULA and DULA for the upper body; and REBA and DEBA scores for full body assessments. The results show that the continuous DULA score follows the RULA score very well in both tasks. The same is true with DEBA and REBA scores. The plots also show the smoothness of the continuous DULA/DEBA scores in comparison to the discrete RULA/REBA. These plots highlight that the majority of DULA/DEBA model errors happen between scores 3 and 4. Figure 8 shows the deviation of the DULA/DEBA scores from RULA/REBA scores for all subjects and tasks. The plot reveals the high accuracy of our learned models. The above results show that DULA and DEBA can be used instead of RULA and REBA in autonomous ergonomics risk assessments.

### C. Postural Optimization

There are three types of information that can be used to correct the human posture: (1) the general rule of thumb of moving toward neutral posture, (2) moving toward the optimal posture calculated from our postural optimization, and (3) use the gradient of DULA/DEBA. The first one is a naive idea that cannot be used in postural optimization in intelligent systems. The optimal posture from our approach is the best one for ergonomists and ergonomically intelligent systems, however it requires solving the postural optimization. In addition to the benefit of gradient of DULA/DEBA in gradient-based postural optimization, it can also provide insight on how to move the posture to increase the human comfort on the fly, since it is fast and does not require solving the postural optimization problem. Figure 9 visualizes the gradient of DULA as arrow markers on the joints and the calculated optimal posture at a the given posture. Larger arrows correspond to relatively higher negative gradient (larger arrow means more increase in human comfort by a change in the joint angle). It shows that at this posture, changes in the trunk and wrist joints have the highest effects on lowering the risk score.

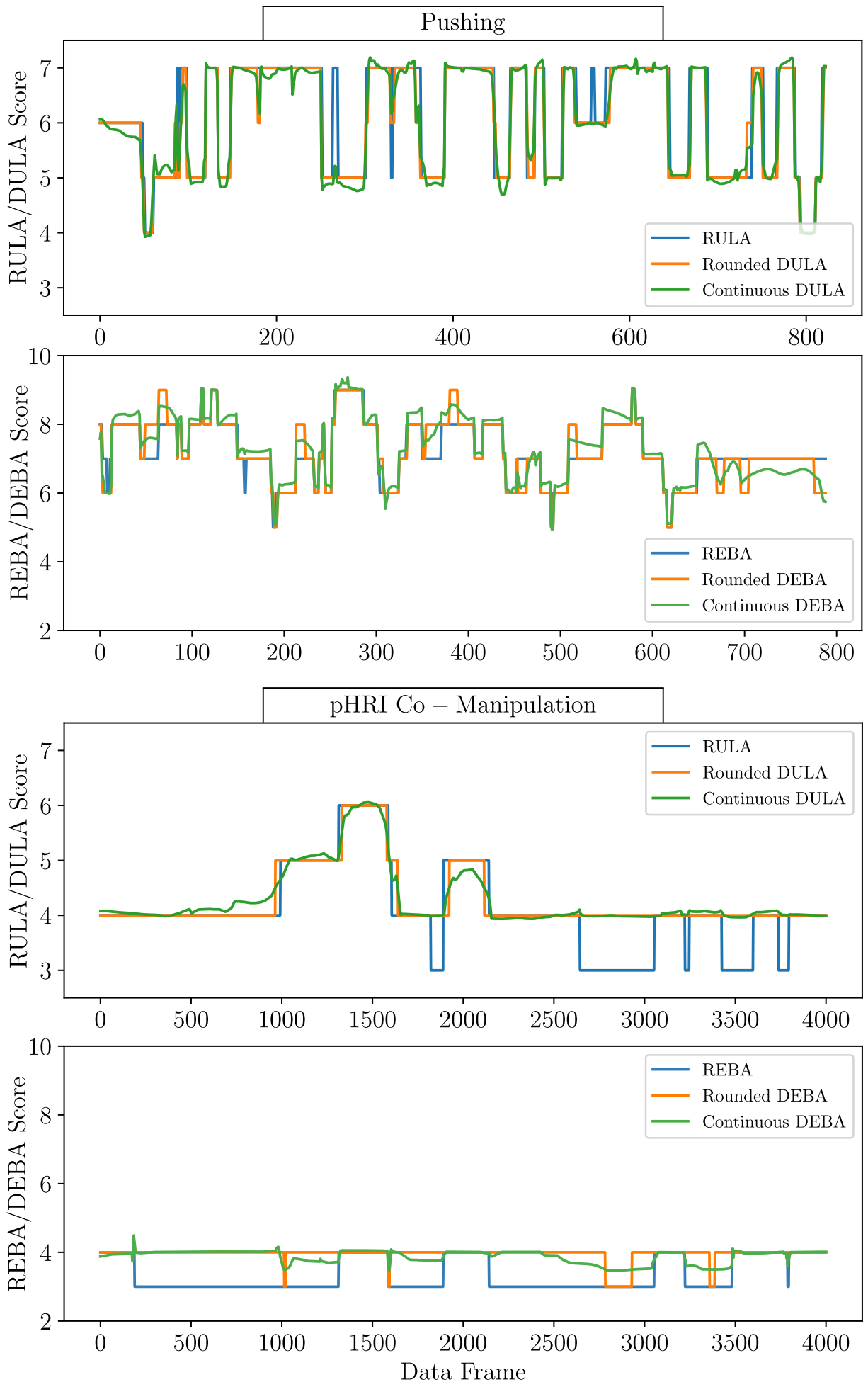

**Fig. 7:** Comparison of (RULA, continuous DULA, and rounded DULA) scores, and (REBA, continuous DEBA, and rounded DEBA) scores for a pushing task (top) and a pHRI co-manipulation with a teleoperated robot (bottom).

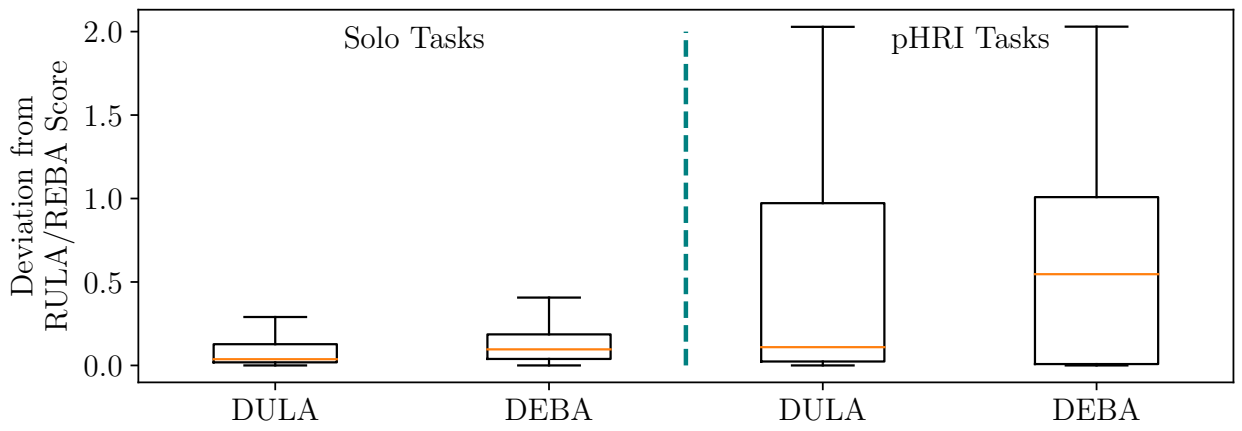

**Fig. 8:** Deviation of rounded DULA and DEBA from RULA and REBA scores among all the subjects and tasks.

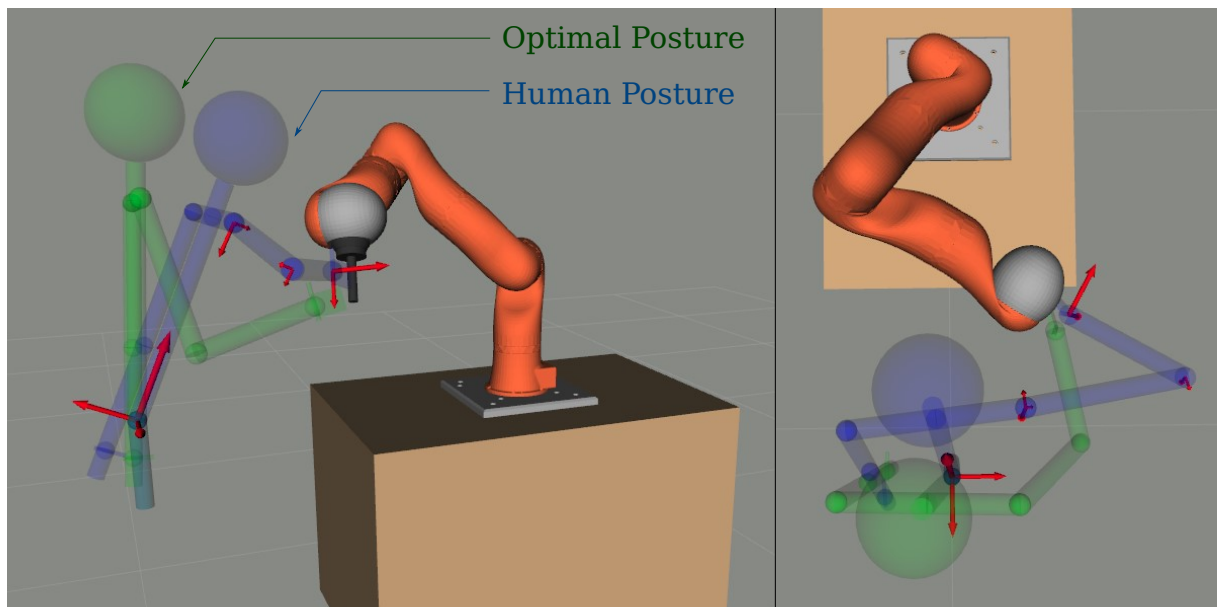

**Fig. 9:** Two views of the optimal posture, and the gradient of DULA at an example posture. We show the scaled gradient of DULA over the human joints using arrow markers. Larger markers correspond to relatively higher values of gradient; the positive angle of joint follows the right hand rule around the markers.

Figure 10 shows the RULA scores for the the postural correction results for gradient-free and gradient-based optimizations in 4 simulated teleoperation tasks with correction acceptance $\alpha \sim \{0, 0.75\}$. The plots show that the risk is reduced after the human applies the suggested optimal postural correction, however, task completion time increases, which is reasonable.

Figure 11 provides more information on comparing the RULA score of resultant optimal postures and the corrected postures (the output of human-like motion generation for a simulated human) using 3 approaches: (a) gradient-free postural optimization with RULA, (b) gradient-free postural optimization with DULA, and (c) gradient-based optimization with DULA for 4 simulated tasks. The whiskers of the box plots are at min and max of the data series. From the left side of the figure, we can see that although all three methods have similar median risk scores for target optimal postures across all tasks, the gradient-free approach using RULA provides optimal postures with lowest risk scores. From the right side, initially we see that the corrected postures of the simulated human are more comfortable after applying the optimal posture from all three methods Moreover, the gradient-free RULA and gradient-based DULA approaches perform similarly and better than the other method. However, the gradient-based approach outperforms the others in terms of much faster computation. Each iteration of the gradient-free approach with RULA using 10000 samples takes ≈ 162s to solve, versus ≈ 0.19s for the gradient-based approach. This makes the gradient-based postural optimization using DULA an ideal approach for postural optimization, especially for online applications. Furthermore, the plot shows that using DULA instead of RULA in a gradient-free optimization degrades the overall performance of postural optimization.

The results from our simple simulated experiments reveal the effect of intelligent postural optimization on lowering the risk of WMSDs in pHRI, and opens an area for future research on performing extensive human subject studies to evaluate postural optimization using DULA and DEBA.

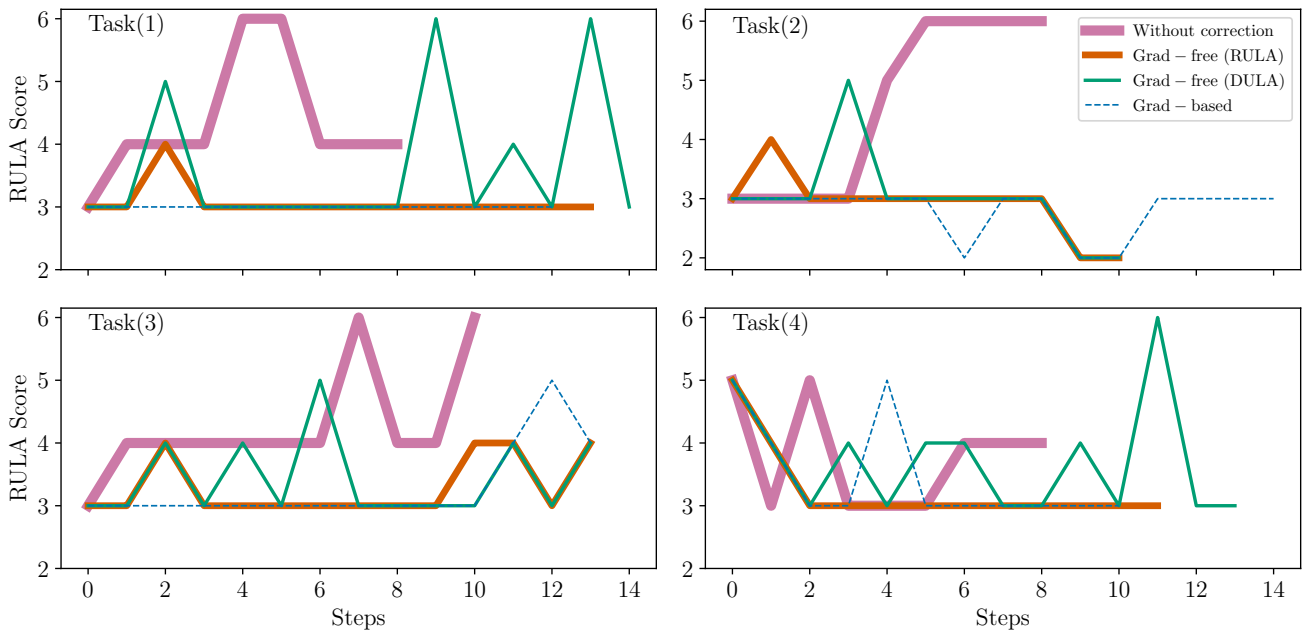

**Fig. 10:** Comparison between the effect of gradient-free and gradient-based postural optimization on 2 teleoperation tasks.

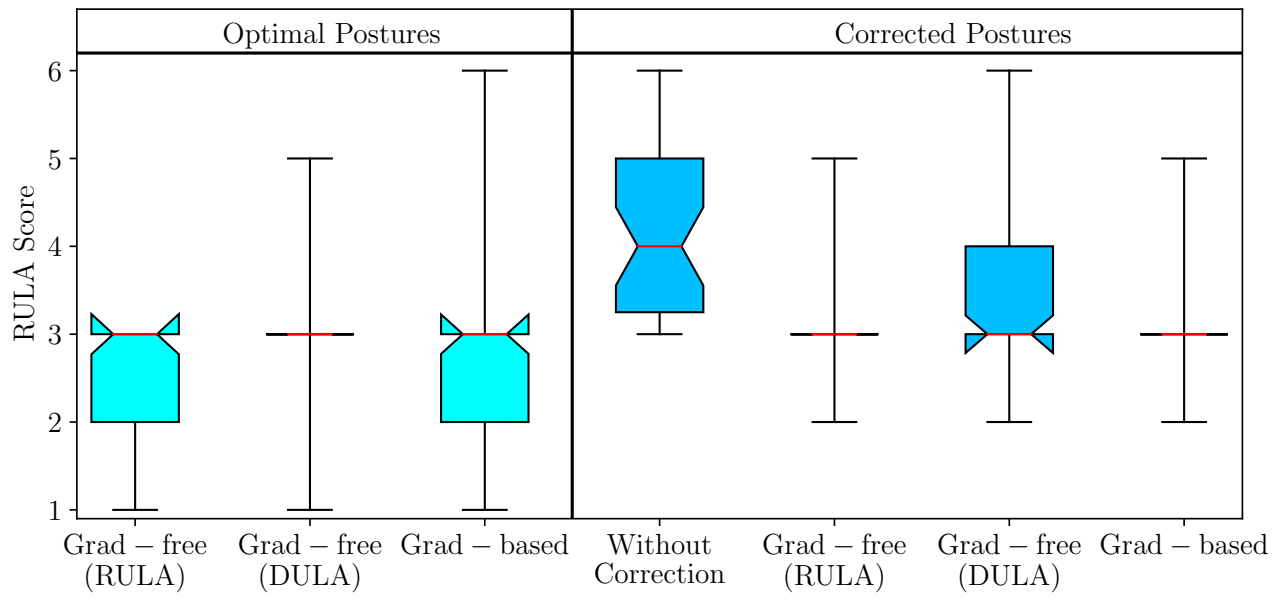

**Fig. 11:** Comparing the RULA score of the optimal postures from gradient-free and gradient-based postural optimizations and their effect on correcting postures in all tasks. Whiskers are at min and max of each data series, and lower score are better.

## VI. Conclusion

This paper proposes a framework for postural optimization in ergonomically intelligent physical human-robot interaction that includes and introduces two differentiable and continuous ergonomics risk assessment models to assess human upper body and full body posture. The DULA and DEBA ergonomics models are learned to replicate the non-differentiable RULA and REBA assessment tool using neural networks. DULA achieves 99.73% and DEBA 99.16% accuracy. The models are designed for computationally efficient postural optimization for pHRI and other related applications. We highlighted the benefits of the differentiable models in gradient-based postural optimization. The results reveal that postural optimizations using DULA and DEBA lower the ergonomics risk score across the tasks, with improved computational performance to using RULA and REBA.

There are several directions for future work. We wish to improve our training using hard negative mining techniques around the region of data with worst performance to further improve the accuracy of our models. The differentiable ergonomics models can be used in human-like motion generation in virtual reality, animation, and games. Our postural optimization work can be immediately extended by conducting a human subject study to evaluate our postural optimization and correction approach. We wish to compare different means of feedback for the posture correction to the human, including visual, auditory, and haptic feedback.

# References

[1] L. Punnett and D. H. Wegman, "Work-related musculoskeletal disorders: the epidemiologic evidence and the debate," *Journal of electromyography and kinesiology*, vol. 14, no. 1, pp. 13–23, 2004.

[2] O. Erdinc and O. Vayvay, "Ergonomics interventions improve quality in manufacturing: a case study," *Intl. Journal of Industrial and Systems Engineering*, vol. 3, no. 6, pp. 727–745, 2008.

[3] W. Yu, T. Ignatius, Z. Li, X. Wang, T. Sun, H. Lin, S. Wan, H. Qiu, and S. Xie, "Work-related injuries and musculoskeletal disorders among factory workers in a major city of china," *Accident Analysis & Prevention*, vol. 48, pp. 457–463, 2012.

[4] D. Gorecky, M. Schmitt, M. Loskyll, and D. Zühlke, "Human-machine-interaction in the industry 4.0 era," in *IEEE international conference on industrial informatics*, 2014, pp. 289–294.

[5] P. K. R. Maddikunta, Q.-V. Pham, B. Prabadevi, N. Deepa, K. Dev, T. R. Gadekallu, R. Ruby, and M. Liyanage, "Industry 5.0: A survey on enabling technologies and potential applications," *Journal of Industrial Information Integration*, p. 100257, 2021.

[6] F. Sherwani, M. M. Asad, and B. Ibrahim, "Collaborative robots and industrial revolution 4.0 (ir 4.0)," in *2020 Intl. Conf. on Emerging Trends in Smart Technologies*. IEEE, 2020, pp. 1–5.

[7] M. Ramaganesh, R. Jayasuriyan, T. Rajpradeesh, S. Bathrinath, and R. Manikandan, "Ergonomics hazard analysis techniques-a technical review," *Materials Today: Proceedings*, 2021.

[8] A. Bazazan, I. Dianat, N. Feizollahi, Z. Mombeini, A. M. Shirazi, and H. I. Castellucci, "Effect of a posture correction–based intervention on musculoskeletal symptoms and fatigue among control room operators," *Applied ergonomics*, vol. 76, pp. 12–19, 2019.

[9] A. A. Shikdar and M. A. Al-Hadhrami, "Smart workstation design: an ergonomics and methods engineering approach," *Intl. Journal of Industrial and Systems Engineering*, vol. 2, no. 4, pp. 363–374, 2007.

[10] H. Motabar and A. D. Nimbarte, "The effect of task rotation on activation and fatigue response of rotator cuff muscles during overhead work," *Applied Ergonomics*, vol. 97, p. 103461, 2021.

[11] A. Yazdani and R. Sabbagh Novin, "Posture estimation and optimization in ergonomically intelligent teleoperation systems," in *Companion of the ACM/IEEE Intl. Conf. on Human-Robot Interaction*, 2021.

[12] A. Yazdani, R. S. Novin, A. Merryweather, and T. Hermans, "Is the leader robot an adequate sensor for posture estimation and ergonomic assessment of a human teleoperator?" in *IEEE Intl. Conf. on Automation Science and Engineering*, 2021.

[13] C. R. Guerrero, J. C. F. Marinero, J. P. Turiel, and V. Muñoz, "Using "human state aware" robots to enhance physical human–robot interaction in a cooperative scenario," *Computer methods and programs in biomedicine*, vol. 112, no. 2, pp. 250–259, 2013.

[14] W. Kim, J. Lee, L. Peternel, N. Tsagarakis, and A. Ajoudani, "Anticipatory robot assistance for the prevention of human static joint overloading in human–robot collaboration," *IEEE Robotics and Automation Letters*, vol. 3, no. 1, pp. 68–75, 2017.

[15] L. Chen, L. F. Figueredo, and M. R. Dogar, "Planning for muscular and peripersonal-space comfort during human-robot forceful collaboration," in *IEEE-RAS Intl. Conf. on Humanoid Robotics*, 2018.

[16] B. Busch, M. Toussaint, and M. Lopes, "Planning ergonomic sequences of actions in human-robot interaction," in *Intl. Conf. on Robotics and Automation*, 2018, pp. 1916–1923.

[17] R. Rahal, G. Matarese, M. Gabiccini, A. Artoni, D. Prattichizzo, P. R. Giordano, and C. Pacchierotti, "Caring about the human operator: haptic shared control for enhanced user comfort in robotic telemanipulation," *IEEE Tran. on Haptics*, vol. 13, no. 1, pp. 197–203, 2020.

[18] L. Peternel, N. Tsagarakis, D. Caldwell, and A. Ajoudani, "Robot adaptation to human physical fatigue in human-robot co-manipulation," *Autonomous Robots*, pp. 1–11, 2018.

[19] S. Chen and R. R. Yang, "Pose trainer: correcting exercise posture using pose estimation," *arXiv preprint arXiv:2006.11718*, 2020.

[20] T. N. Hoang, M. Reinoso, F. Vetere, and E. Tanin, "Onebody: remote posture guidance system using first person view in virtual environment," in *Proceedings of the 9th Nordic Conf. on Human-Computer Interaction*, 2016, pp. 1–10.

[21] L. Peternel, W. Kim, J. Babič, and A. Ajoudani, "Towards ergonomic control of human-robot co-manipulation and handover," in *IEEE-RAS Intl. Conf. on Humanoid Robotics*, 2017, pp. 55–60.

[22] L. Peternel, C. Fang, M. Laghi, A. Bicchi, N. Tsagarakis, and A. Ajoudani, "Human arm posture optimisation in bilateral teleoperation through interface reconfiguration," in *IEEE RAS/EMBS Intl. Conf. for Biomedical Robotics and Biomechatronics*, 2020.

[23] B. Busch, G. Maeda, Y. Mollard, M. Demangeat, and M. Lopes, "Postural optimization for an ergonomic human-robot interaction," in *Intl. Conf. on Intelligent Robots and Systems*, 2017, pp. 2778–2785.

[24] A. G. Marin, M. S. Shourijeh, P. E. Galibarov, M. Damsgaard, L. Fritzsch, and F. Stulp, "Optimizing contextual ergonomics models in human-robot interaction," in *Intl. Conf. on Intelligent Robots and Systems*, 2018, pp. 1–9.

[25] L. Van der Spaa, M. Gienger, T. Bates, and J. Kober, "Predicting and optimizing ergonomics in physical human-robot cooperation tasks," in *Intl. Conf. on Robotics and Automation*. IEEE, 2020, pp. 1799–1805.

[26] S. G. Hart and L. E. Staveland, "Development of nasa-tlx (task load index): Results of empirical and theoretical research," in *Advances in psychology*, vol. 52, 1988, pp. 139–183.

[27] L. McAtamney and E. N. Corlett, "Rula: a survey method for the investigation of work-related upper limb disorders," *Applied ergonomics*, vol. 24, no. 2, pp. 91–99, 1993.

[28] S. Hignett and L. McAtamney, "Rapid entire body assessment (reba)," *Applied ergonomics*, vol. 31, no. 2, pp. 201–205, 2000.

[29] J. Steven Moore and A. Garg, "The strain index: a proposed method to analyze jobs for risk of distal upper extremity disorders," *American Industrial Hygiene Association Journal*, vol. 56, no. 5, 1995.

[30] J. M. Kapellusch, A. Garg, K. T. Hegmann, M. S. Thiese, and E. J. Malloy, "The strain index and acgih tlv for hal: risk of trigger digit in the wistah prospective cohort," *Human factors*, vol. 56, no. 1, 2014.

[31] Z. Khodabakhshi, S. A. Saadatmand, M. Anbarian, and R. Heydari Moghadam, "An ergonomic assessment of musculoskeletal disorders risk among the computer users by rula technique and effects of an eight-week corrective exercises program on reduction of musculoskeletal pain," *Iranian Journal of Ergonomics*, vol. 2, no. 3, 2014.

[32] L. F. Figueredo, R. C. Aguiar, L. Chen, S. Chakrabarty, M. R. Dogar, and A. G. Cohn, "Human comfortability: Integrating ergonomics and muscular-informed metrics for manipulability analysis during human-robot collaboration," *IEEE Robotics and Automation Letters*, vol. 6, no. 2, pp. 351–358, 2020.

[33] A. Shafti, A. Ataka, B. U. Lazpita, A. Shiva, H. A. Wurdemann, and K. Althoefer, "Real-time robot-assisted ergonomics," in *Intl. Conf. on Robotics and Automation*, 2019, pp. 1975–1981.

[34] J. Lim, I. Lee, I. Shim, H. Jung, H. M. Joe, H. Bae, O. Sim, J. Oh, T. Jung, S. Shin, *et al.*, "Robot system of drc-hubo+ and control strategy of team kaist in darpa robotics challenge finals," *Journal of Field Robotics*, vol. 34, no. 4, pp. 802–829, 2017.

[35] A. Rout, B. Deepak, and B. Biswal, "Advances in weld seam tracking techniques for robotic welding: A review," *robotics and computer-integrated manufacturing*, vol. 56, pp. 12–37, 2019.

[36] P. J. D. de Oliveira Evald, R. T. S. da Rosa, J. L. Mór, R. Z. Azzolin, V. M. de Oliveira, and S. S. da Costa Botelho, "Velocity regulation of a linear welding robot by unscented and cubature kalman filter output estimation-based sliding mode control," in *Annual Conf. of the IEEE Industrial Electronics Society*, 2017.

[37] A. Paszke, S. Gross, F. Massa, A. Lerer, J. Bradbury, G. Chanan, T. Killeen, Z. Lin, N. Gimelshein, L. Antiga, A. Desmaison, A. Kopf, E. Yang, Z. DeVito, M. Raison, A. Tejani, S. Chilamkurthy, B. Steiner, L. Fang, J. Bai, and S. Chintala, "Pytorch: An imperative style, high-performance deep learning library," in *Neural Info. Proc. Sys.*, 2019.

[38] F. Pedregosa, G. Varoquaux, A. Gramfort, V. Michel, B. Thirion, O. Grisel, M. Blondel, P. Prettenhofer, R. Weiss, V. Dubourg, J. Vanderplas, A. Passos, D. Cournapeau, M. Brucher, M. Perrot, and E. Duchesnay, "Scikit-learn: Machine learning in Python," *Journal of Machine Learning Research*, vol. 12, pp. 2825–2830, 2011.

[39] M. Tietz, T. J. Fan, D. Nouri, B. Bossan, and skorch Developers, *skorch: A scikit-learn compatible neural network library that wraps PyTorch*, 2017.

[40] L. Vianello, J.-B. Mouret, E. Dalin, A. Aubry, and S. Ivaldi, "Human posture prediction during physical human-robot interaction," *IEEE Robotics and Automation Letters*, vol. 6, no. 3, pp. 6046–6053, 2021.

[41] J. Nocedal and S. Wright, *Numerical optimization*. Springer, 2006.

[42] M. Kobilarov, "Cross-entropy motion planning," *Intl. Journal of Robotics Research*, vol. 31, no. 7, pp. 855–871, 2012.

[43] P. Virtanen, R. Gommers, T. E. Oliphant, M. Haberland, T. Reddy, D. Cournapeau, E. Burovski, P. Peterson, and et al, "SciPy 1.0: Fundamental Algorithms for Scientific Computing in Python," *Nature Methods*, vol. 17, pp. 261–272, 2020.